# A new approach to forecast service parts demand by integrating user preferences into multi-objective optimization

Wenli Ouyang
*Artificial Intelligence Lab, Lenovo Research, Beijing, China.*
Email: ouyangwenli@gmail.com

**Abstract**

Service supply chain management is to prepare spare parts for failed products under warranty. Their goal is to reach agreed service level at the minimum cost. We convert this business problem into a preference based multi-objective optimization problem, where two quality criteria must be simultaneously optimized. One criterion is accuracy of demand forecast and the other is service level. Here we propose a general framework supporting solving preference-based multi-objective optimization problems (MOPs) by multi-gradient descent algorithm (MGDA), which is well suited for training deep neural network. The proposed framework treats agreed service level as a constrained criterion that must be met and generate a Pareto-optimal solution with highest forecasting accuracy. The neural networks used here are two Encoder-Decoder LSTM models: one is used for pre-training phase to learn distributed representation of former generations' service parts consumption data, and the other is used for supervised learning phase to generate forecast quantities of current generations' service parts. Evaluated under the service parts consumption data in Lenovo Group Ltd, the proposed method clearly outperforms baseline methods.

## 1. Introduction

A fundamental aspect of service supply chain management is providing spare parts for replacement to failed products within warranty period. Company who sell products with warranty will have a service level agreement (SLA) with service provider, which defines what level of service is to be provided. Since demand forecast techniques cannot be complete accurate, service provider mitigates the uncertainty of demand forecast by stocking additional quantity as safety stock, in order to achieve agreed service level. This indirect way to estimate spare parts demand may lead to a high chance of overstock or stockout, when setting an appropriate safety stock level involves many subjective human decisions. Overstock results in excessive inventory cost and unhealthy cash flow, while stockout leads to unfulfilled orders and unachievable SLA. Thus, reasonable modeling to simultaneously consider two or more quality criteria: accuracy of demand forecast and service level is a necessity.

Commonly, multi-objective optimization problems (MOPs) can be solved by minimizing a weighted sum of each objective function. However, the linear combination is only valid when multi-objective is not competing with each other, which is rarely the case. An alternative way is finding solutions that are not dominated by any others and said to be Pareto-optimal. A variety of algorithms for MOPs exist. One such approach is the multi-gradient descent algorithm (MGDA) that uses gradient-based optimization to converge to a solution on the Pareto set. It can use the gradients of each objective and solve an optimization problem to decide on an update over parameters, which is well-suited for multi-objective optimization with deep neural networks.

Generally, decision maker (DM) doesn't need a full picture of the Pareto set. A representative subset of the Pareto set or a solution satisfying DM's preferences is acceptable. So-called a prior and a posterior methods are designed to meet these types of needs. One issue often neglected in research literature is the objective-metric mismatch (Huang et al., 2019). The domain knowledge of DM is at metric space, rather than at objective space. Most of the time, metrics of interest such as service level can't be optimized directly, which are non-decomposable (involving statistics of a set of examples). Differentiable objective functions need be designed to be good approximations to metrics. Here we propose a general framework that includes a posterior and a prior methods. The general framework integrates DM's preferences in metric space into MOPs, uses MGDA to optimize till converging to a point on the Pareto set and produces DM preferred Pareto optimal solutions. Instead of coming up with a Pareto-optimal solution without preferences, the proposed framework can treat agreed service level as a constrained criterion that must be met and generate Pareto-optimal solutions with highest forecasting accuracy. We empirically evaluate the presented methods on service parts consumption data in Lenovo Group Ltd and show our method clearly outperforms baseline methods.

## 2. Related Work

**Preference based multi-objective optimization**. A good summary of preference based multi-objective optimization has been written by Wang et al. (2017). According to the time when preference is articulated, we can classify preference based multi-objective optimization into three categories: a priori method, a posterior method and interactive method (Branke, Jürgen; Deb, Kalyanmoy; Miettinen, Kaisa; Słowinski, 2008). A priori method needs sufficient preference information before the solution process, such as utility function method (Jeantet & Spanjaard, 2011; Keeney & Raiffa, 1976) and lexicographic ordering method (Fishburn, 2008; Kaufman & Michalski, 1999; Pinchera, Perna, & Migliore, 2017). A posteriori method aims at producing a representative subset of the Pareto optimal solutions. Normal boundary intersection (Indraneel Das & Dennis, 2003; De Motta, Afonso, & Lyra, 2012) and normal constraint methods (A. Messac, Ismail-Yahaya, & Mattson, 2003) are well-known examples. Most existing multi-objective evolutionary algorithms (MOEAs) belong to this category (Zitzler, 2012). Interactive method involves DM in optimization process. The DM continuously interacts with the optimization method when searching for the preferred solution (Ben Said, Bechikh, & Ghedira, 2010; Miettinen & Ruiz, 2016). Interactive methods have various ways to articulate preference information, such as reference vectors (Chugh, Jin, Miettinen, Hakanen, & Sindhya, 2018; Deb & Kumar, 2007; Deb & Sundar, 2006; Luque, Miettinen, Eskelinen, & Ruiz, 2009) and preference relation (Battiti & Passerini, 2010; Branke, Jürgen; Deb, Kalyanmoy; Miettinen, Kaisa; Słowinski, 2008; Deb, Sinha, Korhonen, & Wallenius, 2010). In our work, we present a posterior method able to generate a representative subset of Pareto frontier efficiently and a prior method that DM can specify criteria to find a preferred optimal solution.

**Service parts demand forecast**. The research literature on demand forecast is enormous. We summarized the work most closely related to service parts and refer the interested reader to reviews by Bacchetti and Saccani (2012) and Boylan and Syntetos (2010). Service parts demand patterns can be classified into two categories: intermittent and non-intermittent based on the mean inter-demand interval. The most widely studied and applied intermittent demand forecast method is Croston method and its variants (Croston, 1972; Teunter, Syntetos, & Babai, 2011; Willemain, Smart, Shockor, & DeSautels, 1994). Several other methods to forecast intermittent demand have appeared, such as aggregation-based methods (Babai, Ali, & Nikolopoulos, 2012), bootstrapping-based methods (Porras & Dekker, 2008; Syntetos, Zied Babai, & Gardner, 2015; Willemain, Smart, & Schwarz, 2004) and neural networks (Gutierrez, Solis, & Mukhopadhyay, 2008; Kourentzes, 2013). Non-intermittent demand forecast is more correlated with the installed base of products (Van der Auweraer, Boute, & Syntetos, 2019). Installed base is defined as the number of sold products that can lead to demand for spare parts (Kim, Dekker, & Heij, 2017). Dekker et al. (2013) argue that the demand of spare parts follows the demand for the installed product with a delay and has distinct patterns in three phases of life cycle: growth, maturity, decline. Jin and Tian (2012) proposed a method to estimate demand of service parts under the time-varying installed base in the growth phase. Chou et al. (2015, 2016) studied the demand pattern of service parts in decline phase till end of life cycle and estimate the overstocking cost as unused service parts will become obsolete. Kim et al. (2017) presented a general methodology to forecast end-of-life spare part demand based on four types of installed base.

## 3. Multi-Objective Optimization Method

Consider MOPs as over an input space $X$, a output space $Y$ and a collection of objective function spaces $\boldsymbol{L} = \{L^t\}_{t \in [T]}$ and metric spaces $\boldsymbol{M} = \{M^t\}_{t \in [T]}$, such that a dataset of data points $\{x_i, y_i\}_{i \in [N]}$ is given where $T$ is the number of objectives, $N$ is the number of data points, and $x_i$ and $y_i$ are the $i^{th}$ data point. We further consider a parametric hypothesis class as $f(x; \boldsymbol{\theta}): X \to Y$, objective-specific loss functions as $L^t(\cdot, \cdot): Y \times Y \to \mathbb{R}^+$ and corresponding metric functions as $M^t(\cdot, \cdot): Y \times Y \to \mathbb{R}^+$. Generally, the empirical risk minimization formulation is written as:

$$\min_{\boldsymbol{\theta}} \sum_{t=1}^{T} c^t \hat{L}^t(\boldsymbol{\theta})$$

where $\hat{L}^t(\boldsymbol{\theta})$ is the empirical loss of the objective $t$, defined as $\hat{L}^t(\boldsymbol{\theta}) \triangleq \frac{1}{N}\sum_i L^t(f(x_i; \boldsymbol{\theta}), y_i)$. This linear (weighted) combination of the objective functions has many drawbacks. User often specifies preferences (criteria) in metric space $\boldsymbol{M}$ and is uncertain how to set weighting coefficients in objective space $\boldsymbol{L}$. Furthermore, it is well known that it can generate only the convex part of a Pareto frontier (I. Das & Dennis, 1997; Koski, 1985; Achille Messac, Sundararaj, Tappeta, & Renaud, 2012) while real-life problems often results in non-convex Pareto frontiers. Alternatively, MOPs can be formulated as multi-objective optimization, optimizing a collection of conflicting objectives in $L$, defined as:

$$\min_{\theta} L(\theta) = \lim_{\theta} \left( \hat{L}^1(\theta), \dots, \hat{L}^t(\theta) \right)^T$$

The goal of multi-objective optimization is achieving Pareto optimality, defined as:
Definition 1 (Pareto optimality for MOPs)

(a) A solution $\theta$ dominates a solution $\bar{\theta}$ if $\hat{L}^t(\theta) \leq \hat{L}^t(\bar{\theta})$ for all objectives $t$ and $L(\theta) \neq L(\bar{\theta})$.

(b) A solution $\theta^*$ is called Pareto optimal if there exists no solution $\theta$ that dominates $\theta^*$.

The set of Pareto optimal solutions is called the Pareto set ($P_\theta$). Its image is called the Pareto frontier ($P_L = \{L(\theta)\}_{\theta \in P_\theta}$) in objective space $L$ and ($P_M = \{M(\theta)\}_{\theta \in P_\theta}$) in metric space $M$. In this paper, we focus on gradient-based multi-objective optimization due to its compatibility with deep neural networks.

In the rest of this section, we first summarize in Section 3.1 how multi-objective optimization can be performed with gradient descent and its practical adaptation to learning over very large parameter spaces of deep neural network. Then, in section 3.2 we propose a posterior method to obtain Pareto frontier efficiently. Finally, in section 3.3 we propose a prior method to generate user preferred solution.

### 3.1 Multiple Gradient Descent Algorithm

In this section, we summarize the multiple gradient descent algorithm (Désidéri, 2012; Sener, 2018). MGDA leverages the Karush-Kuhn-Tucker (KKT) conditions, which are necessary for optimality (Désidéri, 2012; Fliege & Svaiter, 2000). The KKT conditions is defined as:

There exist $\alpha^1, \dots, \alpha^T \geq 0$ such that $\sum_1^T \alpha^t = 1$ and $\sum_1^T \alpha^t \nabla_\theta \hat{L}^t(\theta) = 0$.

Any solution that satisfies these conditions is called a Pareto stationary point. Consider the optimization problem:

$$\min_{\alpha^1, \dots, \alpha^T} \left\{ \left\| \sum_{t=1}^T \alpha^t \nabla_\theta \hat{L}^t(\theta) \right\|_2^2 \,\middle|\, \sum_1^T \alpha^t = 1, \alpha^t \geq 0 \,\forall t \right\}$$

The solution ($\sum_{t=1}^T \alpha^t \nabla_\theta$) is either 0 satisfying the KKT conditions or gives a descent direction improving all objectives (Désidéri, 2012). The optimization problem defined above has been studied widely in computational geometry. Their case has a large number of points in low dimensional space that is in contrary to our case, training deep neural networks that have a small number of points (objectives) and a large number of spaces (parameters). Sener (2018) proposed a Frank-Wolfe solver that able to tackle the problem well. The detail of this MGDA is given in Algorithm 1.

### 3.2 A posterior method

The algorithm proposed by Sener (2018) is perfect for deep neural networks with multiple objectives. However, it only produces one single solution in the Pareto frontier. Here, we propose a posterior method that able to generate a representative subset of Pareto frontier.

The basic idea is that we can change $\boldsymbol{\alpha}$ at each iteration by integrating with a subjective weights vector $\boldsymbol{W} = (w^1, \ldots, w^T)$. The revised $\boldsymbol{\alpha}$ is defined as $\left\{\alpha^t = \frac{\alpha^t w^t}{\sum_1^T \alpha^t w^t}\right\}_{t \in [T]}$. The proof by Désidéri (2012) is still hold for new $\boldsymbol{\alpha}$ to guarantee a Pareto stationary point. An algorithm capable to obtain a representative subset of the Pareto frontier ($P_M = \{M(\theta)\}_{\theta \in P_\theta}$) in metric space $M$ is necessary for DM to make appropriate decisions. Because the knowledge decision maker (DM) has is in the metric space $M$ rather than in the objective space $L$. We express DM's knowledge as $\{\alpha^t \leq M^t \leq \beta^t\}_{t \in [T]}$.[1] The most crucial capability of the posterior method is to explore the Pareto frontier efficiently. The exploring strategy used here is a heuristic method that search the most unknown region of the most unexplored metric $t$. This greedy method makes the optimal choice at each step and has an adjustable parameter $\eta$, controlling the aggressiveness of the searching pace. Another user-specified parameter is a threshold vector $\boldsymbol{\varphi} = \{\varphi^t\}_{t \in [T]}$ to regulate the granularity of the desired representative Pareto frontier. At each exploring step, the posterior method keeps track of the achieved metric points as $\{M_1, \ldots, M_k, \ldots\}$. Add $\{\alpha^t, \beta^t\}_{t \in [T]}$ to $\{M_1, \ldots, M_k, \ldots\}$, sort it increasingly and get $\{\alpha^t, M_m^t, \ldots, M_k^t, \ldots, M_n^t, \beta^t\}_{t \in [T]}$. Granularity for metric $t$ is defined as averaged gap between adjacent elements in sequence $\{\alpha^t, M_m^t, \ldots, M_k^t, \ldots, M_n^t, \beta^t\}$. The posterior algorithm is given in Algorithm 2.

### 3.3 A prior method

The most common way of solving MOPs is by priori articulation of the DM's preferences. Instead of producing a subset of Pareto optimal solutions, the prior method produces a single solution satisfying DM's preferences. Here we consider the case that the DM's preferences are articulated as hard constraints. The hard constraints in the metric spaces $M$ is written as:

$$g_i(M) = c_i \text{ for } i = 1, \ldots, n \text{ and } h_j(M) \geq b_j \text{ for } j = 1, \ldots, m$$

The general case is too complicated to solve in a universal way and unnecessary in most of the circumstances. We only consider the scenario that each constraint is only about one metric variable $M^t$, The simplified one is written as:

$$g_i(M^i) = c_i \text{ for } i = 1, \ldots, n \text{ and } h_j(M^j) \geq b_j \text{ for } j = 1, \ldots, m \text{ and } n, m \leq T$$

We assume the equations and inequalities are solvable and the solution for metric $t$ is expressed as:

$$\boldsymbol{cond}^t = (cond_1^t := M^t \geq A_t \ \lor\ cond_2^t := M^t \leq B_t \ \lor\ cond_3^t := M^t = C_t \ \lor\ cond_4^t := D_t \geq M^t \geq E_t \ \ldots \lor\ cond_{N_t}^t := \cdots)$$

The prior method proposed here is designed to search a Pareto optimal solution in the feasible set that satisfies $\overline{\boldsymbol{Cond}} = (\boldsymbol{cond}^1 \land \ldots \land \boldsymbol{cond}^k | k \leq T)$. The feasible set can be partitioned into feasible subsets such as $\boldsymbol{S}^1 = (cond_1^1 \land \ldots \land cond_1^k)$. The extreme points of this subset in space $\{M^t\}_{t \in [k]}$ is $\boldsymbol{P}^1 = (A_1, \ldots, A_k)$. We define the Euclidean distance between a metric point $\boldsymbol{M}$ in space $\{M^t\}_{t \in [T]}$ and the extreme point $\boldsymbol{P}^1$ in space $\{M^t\}_{t \in [k]}$ as $\sqrt{(M^1 - A_1)^2 + \cdots + (M^k - A_k)^2}$. The prior method prioritizes the searching process from the feasible subset having largest number of inequalities and smallest

Euclidean distance to the one having largest number of equations and largest Euclidean distance. The prior method is given in Algorithm 3.

---

**Algorithm 1**: MGDA for MOPs

---

1: $\alpha^1, \ldots, \alpha^T$ = FRANKWOLFESOLVER($\boldsymbol{\theta}$)
2: $\boldsymbol{\theta} = \boldsymbol{\theta} - \eta \sum_1^T \alpha^t \nabla_{\boldsymbol{\theta}} \hat{L}^t(\boldsymbol{\theta})$

3: **procedure** FRANKWOLFESOLVER($\boldsymbol{\theta}$)
4:   Initialize $\boldsymbol{\alpha} = (\alpha^1, \ldots, \alpha^T) = \left(\frac{1}{T}, \ldots, \frac{1}{T}\right)$
5:   Precompute M st. $M_{i,j} = \left(\nabla_{\boldsymbol{\theta}} \hat{L}^i(\boldsymbol{\theta})\right)^T \left(\nabla_{\boldsymbol{\theta}} \hat{L}^i(\boldsymbol{\theta})\right)$
6:   **repeat**
7:     $\hat{t} = \text{argmin}_r \sum_t \alpha^t M_{rt}$
8:     $\hat{\gamma} = \text{argmin}_\gamma \left((1-\gamma)\boldsymbol{\alpha} + \gamma \mathbf{e}_{\hat{t}}\right)^T M\left((1-\gamma)\boldsymbol{\alpha} + \gamma \mathbf{e}_{\hat{t}}\right)$
9:     $\boldsymbol{\alpha} = (1-\hat{\gamma})\boldsymbol{\alpha} + \hat{\gamma}\widehat{\mathbf{e}_{\hat{t}}}$
10:  **until** $\hat{\gamma} \sim 0$ **or** Number of Iterations Limit
11:  **return** $\alpha^1, \ldots, \alpha^T$
12: **end procedure**

---

**Algorithm 2**: A Posterior Method

---

1: Initialize $\boldsymbol{W} = (w^1, \ldots, w^T) = \left(\frac{1}{T}, \ldots, \frac{1}{T}\right)$ and $k = 1$
2: **repeat**
3:   $M^1, \ldots, M^t$ = OPTIMIZE($\boldsymbol{W}$)
4:   Memorize weights $\boldsymbol{W}$ and metrics $\boldsymbol{M}$ as $\boldsymbol{W}_k$ and $\boldsymbol{M}_k$ in List **T**
5:   $\boldsymbol{W}$ = REWEIGHTING1(**T**)
6:   $k = k + 1$
7: **until** $\boldsymbol{W}$ is null **or** Number of Iteration Limit
8: Metrics vectors $\{\boldsymbol{M}_1, \ldots, \boldsymbol{M}_k, \ldots\}$ in List **T** is the representative subset of Pareto frontier

9: **procedure** OPTIMIZE($\boldsymbol{W}$)
10:  **repeat**
11:    $\alpha^1, \ldots, \alpha^T$ = FRANKWOLFESOLVER($\boldsymbol{\theta}$)  ▷ Same as the one in Algorithm 1
12:    reweighting $\boldsymbol{\alpha}$ st. $\alpha^t = \frac{\alpha^t w^t}{(\sum_1^T [(\alpha^t w)^t])}$
13:    $\boldsymbol{\theta} = \boldsymbol{\theta} - \eta \sum_1^T \alpha^t \nabla_{\boldsymbol{\theta}} \hat{L}^t(\boldsymbol{\theta})$
14:    $\boldsymbol{M}$ st. $M^t(\boldsymbol{\theta}) \triangleq M^t(f(X; \boldsymbol{\theta}), Y)$  ▷ Monitor the progress of metrics
15:  **until** $\boldsymbol{M}$ Stop Improving **or** Number of Iteration Limit
16:  **return** $\boldsymbol{M}$
17: **end procedure**

18: **procedure** REWEIGHTING1(**T**)
19:  Calculate the granularity of obtained Pareto frontier in the range $\{\alpha^t \leq M^t \leq \beta^t\}_{t \in [T]}$ for each metric as $\widehat{\boldsymbol{\varphi}} = \{\widehat{\varphi}^t\}$
20:  **if** $\widehat{\boldsymbol{\varphi}} \leq \boldsymbol{\varphi}$
21:    $\boldsymbol{W}$ = null
22:  **else**
23:    $\hat{t} = \text{argmax}_t\{\widehat{\varphi}^t\}_{t \in [T]}$  ▷ select the metric of highest granularity
24:    Find two adjacent point $M_i^{\hat{t}}$ and $M_j^{\hat{t}}$ in sorted sequence $\{\alpha^{\hat{t}}, M_m^{\hat{t}}, \ldots, M_k^{\hat{t}}, \ldots, M_n^{\hat{t}}, \beta^{\hat{t}}\}$ that have largest interval

25:       **if** $M_i^{\hat{t}}$ equals $\alpha^{\hat{t}}$

26:            $\boldsymbol{W} = (w^1, w^2, \dots, w^T)$ and $w^t = \begin{cases} w_n^t & \text{if } t \neq \hat{t} \\ \frac{1}{\eta} w_n^t & \text{if } t = \hat{t} \end{cases}$

27:       **else if** $M_j^{\hat{t}}$ equals $\beta^{\hat{t}}$

28:            $\boldsymbol{W} = (w^1, w^2, \dots, w^T)$ and $w^t = \begin{cases} w_n^t & \text{if } t \neq \hat{t} \\ \eta w_n^t & \text{if } t = \hat{t} \end{cases}$

29:       **else**

30:            $\boldsymbol{W} = \frac{W_i + W_j}{2}$    ▷ half between the corresponding weights $\boldsymbol{W}_i$ and $\boldsymbol{W}_j$

31:   **return** $\boldsymbol{W}$

32: **end procedure**

---

**Algorithm 3**: A Prior Method

---

1: Initialize $\boldsymbol{W} = (w^1, \dots, w^T) = \left(\frac{1}{T}, \dots, \frac{1}{T}\right)$

2: $M^1, \dots, M^t = \text{OPTIMIZE}(\boldsymbol{W})$    ▷ Same as the one in Algorithm 2

3: **if** $\boldsymbol{M}$ satisfies the conditions ($\boldsymbol{cond}^1 \wedge \dots \wedge \boldsymbol{cond}^k | k \leq T$)

4:   **return** $\boldsymbol{M}$ to DM

5: **else**

6:   **enumerate** the feasible subsets $\boldsymbol{S}$ of ($\boldsymbol{cond}^1 \wedge \dots \wedge \boldsymbol{cond}^k | k \leq T$) by priority

7:       find the extreme point $\boldsymbol{P}$ of the feasible subset $\boldsymbol{S}$

8:       **repeat**

9:           $\boldsymbol{W} = \text{REWEIGHTING2}(\boldsymbol{P}, \boldsymbol{M}, \boldsymbol{W}, \boldsymbol{S})$

10:          $M^1, \dots, M^t = \text{OPTIMIZE}(\boldsymbol{W})$    ▷ Same as the one in Algorithm 2

11:      **until** $\boldsymbol{M}$ is in the subset $\boldsymbol{S}$ **or** Number of Iteration Limit

12:      **break if** $\boldsymbol{M}$ is in the subset $\boldsymbol{S}$

13:   **end enumerate**

14:   **return** $\boldsymbol{M}$ to DM

15: **procedure** REWEIGHTING2($\boldsymbol{P}, \boldsymbol{M}, \boldsymbol{W}, \boldsymbol{S}$)

16:   find the metric set $U = \{i, \dots, j\}_{i,j \in k}$ that don't satisfy corresponding constraints in $\boldsymbol{S}$

17:   $\hat{t} = \text{argmax}_t \{\|P_t - M_t\|\}_{t \in U}$    ▷ select the metric of highest gap to extreme point in $U$

18:   **if** $M_{\hat{t}} > P_{\hat{t}}$

19:       $\boldsymbol{W} = (w^1, w^2, \dots, w^T)$ and $w^t = \begin{cases} w_n^t & \text{if } t \neq \hat{t} \\ \frac{1}{\eta} w_n^t & \text{if } t = \hat{t} \end{cases}$

20:   **else**

21:       $\boldsymbol{W} = (w^1, w^2, \dots, w^T)$ and $w^t = \begin{cases} w_n^t & \text{if } t \neq \hat{t} \\ \eta w_n^t & \text{if } t = \hat{t} \end{cases}$

22:   **return** $\boldsymbol{W}$

23: **end procedure**

## 4. Service Parts Predictive Model

In this paper, we propose a long short-term memory (LSTM) based architecture consisting of two Encoder-Decoder LSTM networks to forecast service parts demand. The LSTM model gained popularity recently (Cho et al., 2014; Donahue et al., 2017; Graves, Mohamed, & Hinton, 2013; Hochreiter & Schmidhuber, 1997; Srivastava, 2015; Sutskever, Vinyals, & Le, 2014; Zhu & Laptev, 2017), due to its capability to model end-to-end nonlinear interactions, incorporate exogenous variables and extract features automatically (Assaad, Boné, & Cardot, 2008). By using LSTM models, we aim to develop a versatility, being able to forecast both non-intermittent and intermittent demand at all phases of life cycle. The complete architecture is shown in Figure 1. Prior to training the prediction model, we fit an encoder-decoder that can extract representative embedding from former generations' consumption data. By using the first $k$ timestamps to predicting the following $g$ timestamps, the embedding state of the encoder must extract representative features of former generations' consumption data by constructing a fixed-dimensional embedding state. This learned embedding is helpful for forecasting in the transitional phases of life cycle like growth and decline, because consumptions of former generation's parts have similar transitional patterns caused by regular customer behaviors. After the pre-training of the first encoder-decoder, the last state of the encoder is treated as learned embedding of former generations' parts and used for supervised learning of current generation parts. As there are external and internal data of current generation parts available, they can be concatenated with the embedding feature and passed together to the second encoder-decoder network. After the full model is trained, the inference stage involves only the encoder of the first encoder-encoder network for extracting learned embedding of former generations' parts and the whole second encoder-encoder network for predicting further timestamps of current generation parts.

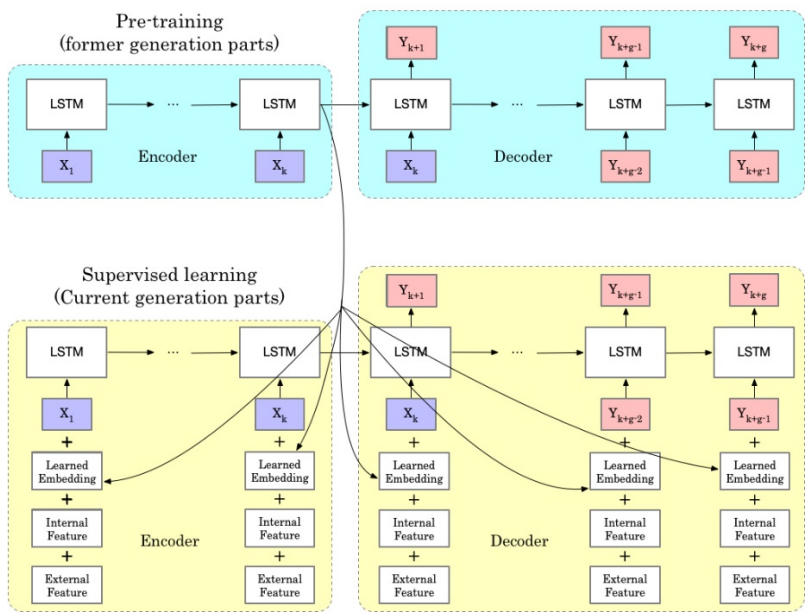

Figure 1. Illustration of the proposed framework

## 5. Implementation Details

The proposed multi-objective optimization method is evaluated by using service parts consumption data in Lenovo Group Ltd over five years across three regions, including Europe, the Middle East and Africa (EMEA), Asia Pacific and Japan (AP) and North America (NA). Except service parts consumption data, other internal data like installed base, service parts category, etc., and external data like weather condition and holiday are collected. The full feature we used is listed in Table 1. All the data is aggregated by week. We reduce the original dimensionality of categorical data by taking the 4$^{th}$ root of the number of categories. We use four years of data as the training set, the following six months as the validation set, and the final six months as the testing set. For both pre-training and supervised learning, we encode the previous 26 weeks features as input and predict the upcoming 26 weeks consumption quantities. The first encoder-decoder network is constructed with two-layer LSTM, with 84 and 32 hidden states, respectively. The second encoder-decoder network is also a two-layer LSTM, with 512 and 128 hidden states, respectively. In order to training the networks to be capable to generate long-term forecast, teacher forcing that use the real target outputs as next input is only used at the beginning of the training and gradually convert to use decoder's guess as the next input.

| type | dimension | Feature |
| --- | --- | --- |
| Internal | 26 | Log-transformed consumption data |
| Internal | 84 | Leaned embedding of former generations' parts consumption data |
| Internal | 26 | Installed base data |
| Internal | 1 | Number of consecutive weeks have no consumption |
| Internal | 4 | Service parts categories |
| Internal | 1 | Region type |
| External | 3 | Week of the year |
| External | 2 | Month of the year |
| External | 1 | Season of the year |
| External | 1 | Local holiday |
| External | 1 | Local humidity |
| External | 1 | Local temperature |

Table 1. Summarization of features

To measure the effectively of the proposed multi-objective optimization method, we adopt two evaluation metrics used in service supply chain, namely accuracy $ACC$ and service level $SL$. Mathematically, they are defined as follows:

$$ACC = \sum_{n=1}^{N} \frac{\sum_{t=k+1}^{k+g} y_t \min(\sum_{t=k+1}^{k+g} y_t, \sum_{t=k+1}^{k+g} \hat{y}_t)}{\sum_{n=1}^{N} \sum_{t=k+1}^{k+g} y_t \max(\sum_{t=k+1}^{k+g} y_t, \sum_{t=k+1}^{k+g} \hat{y}_t)}$$

$$SL = \sum_{n=1}^{N} \frac{\sum_{t=k+1}^{k+g} y_t}{\sum_{n=1}^{N} \sum_{t=k+1}^{k+g} y_t} \min(\frac{\sum_{t=k+1}^{k+g} \hat{y}_t}{\sum_{t=k+1}^{k+g} y_t}, 1)$$

where $\hat{y}_t$ and $y_t$ denote predicted and real quantities respectively. Because these two metrics are non-decomposable and can't be optimized directly. We use mean square error ($MSE$) and quantile regression loss ($QRL$) as loss functions for optimization. They are defined as follows:

$$MSE = \sum_{n=1}^{N} \sum_{t=k+1}^{k+g} (y_t - \hat{y}_t)^2$$

$$QRL = \sum_{n=1}^{N} \sum_{t=k+1}^{k+g} \max(q(y_t - \hat{y}_t), (q-1)(y_t - \hat{y}_t))$$

where $q \in (0,1)$ is quantile parameter and is set to be 0.9 to prefer overstock rather than stockout. The baseline methods we consider are (i) **static scaling posterior method:** replace MGDA in the posterior method to linear combination of loss functions $\sum_{t=1}^{T} c^t L^t$ with static scaling from $\{c^t \in [0,1] | \sum_t c^t = 1\}$, (ii) **static scaling prior method:** replace MGDA in the prior method to the same method as (i), (iii) **grid search:** exhaustively trying various values from $\{c^t \in [0,1] | \sum_t c^t = 1\}$ and optimizing the linear combination of loss functions $\sum_{t=1}^{T} c^t L^t$.

## 6. Results

Comparing the proposed posterior method with the baseline method (i), The Pareto frontier generated by our method clearly has better solutions than the one generated by baseline method (i). We visualize two Pareto frontiers in Figure 2 (a). It demonstrates the necessity of using MGDA to solve MOPs, whose weights for each objective is adaptive rather than static at each optimization step. Compared with the baseline methods (iii) whose majority of efforts is working on a portion of the Pareto frontier, our method contributes the efforts evenly to find the solutions on the Pareto frontier. Under the same number of runs, it covers two times bigger frontier than the baseline method (iii), as shown in Figure 2(b). Most likely, DM will set a threshold for certain metric such as **SL** in our case. The proposed prior method is set to meet DM's threshold of **SL** and give optimal results for the other metric

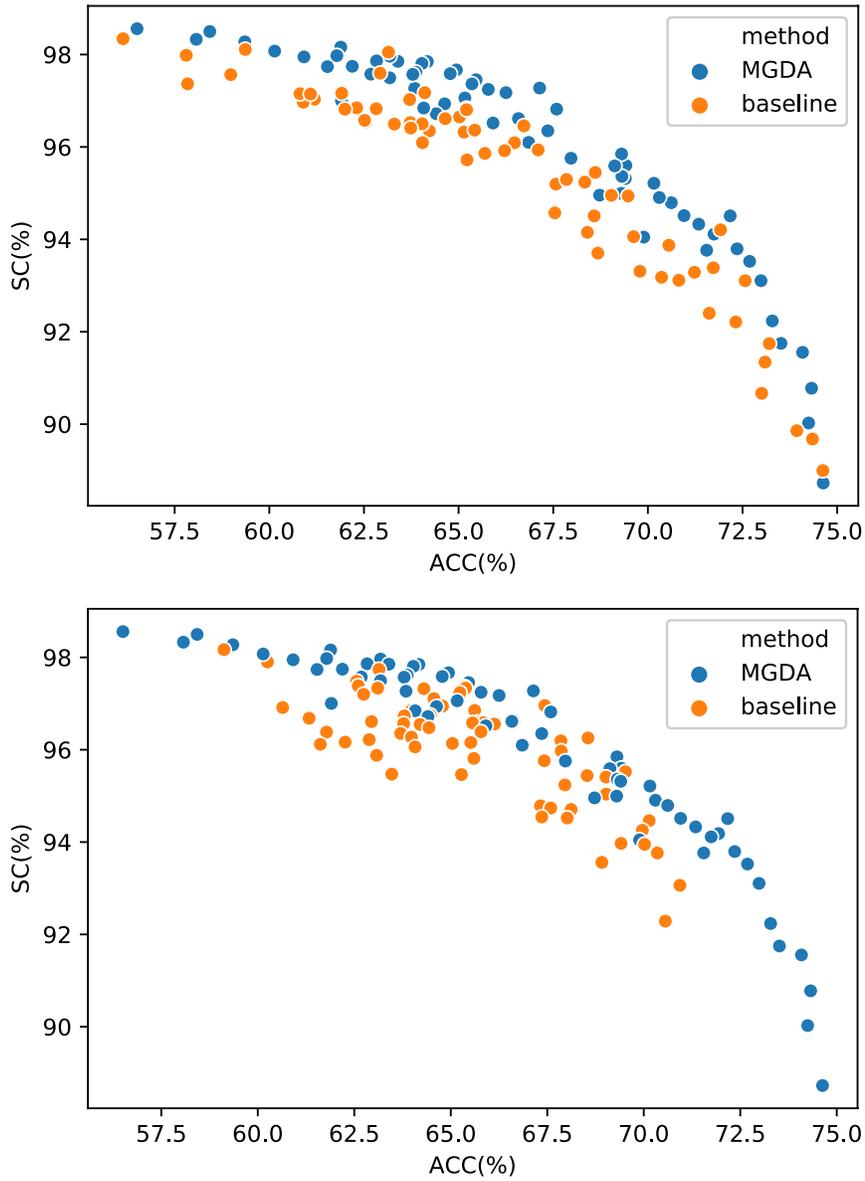

Figure 1. the Pareto frontiers comparison (a) between our method with baseline method (i); (b) between our method with baseline method (iii).

***ACC***. Compared with the baseline method (ii), our method finds solutions that has higher ***ACC*** when different threshold of ***SL*** is satisfied. The results are listed in Table 2.

| Threshold of ***SL*** | ***ACC*** of our method | ***ACC*** of baseline method (ii) |
|---|---|---|
| 98% | 61.3 | 60.1 |
| 95% | 72.5 | 70.4 |

| Threshold of *SL* | *ACC* of our method | *ACC* of baseline method (ii) |
|---|---|---|
| 92% | 73.8 | 72.6 |
| 90% | 74.9 | 74.3 |

Table 2. Comparison of the prior methods with MGDA optimizer and linear combination optimizer.

## 7. Conclusion

We have proposed a preference based multi-objective optimization framework, supporting both a posterior and a prior methods. Applied in the service parts demand forecasting scenario, the proposed posterior method generated the Pareto frontier more efficiently and produced the solutions that have higher accuracy and higher service level, and the proposed prior method produced solutions with DM satisfied service level and higher accuracy. Our framework is capable to be applied to any neural network with multiple objectives without any modifications.